\pdfoutput=1
\PassOptionsToPackage{table,xcdraw}{xcolor}
\documentclass[11pt]{article}

\usepackage[final]{coling}

\usepackage{times}
\usepackage{latexsym}
\usepackage{tipa}

\usepackage[T1]{fontenc}
\usepackage{comment}
\usepackage[utf8]{inputenc}

\usepackage{microtype}

\usepackage{hyperref}

\usepackage{inconsolata}

\usepackage{graphicx}

\usepackage{float}
\usepackage[table,xcdraw]{xcolor}
\usepackage{dashundergaps}

\usepackage{multirow}
\usepackage{makecell} 

\title{A Benchmark of French ASR Systems Based on Error Severity}

\author{Antoine Tholly$^1$, Jane Wottawa$^2$, Mickael Rouvier$^3$, Richard Dufour$^1$ \\
  $^1$LS2N, Nantes Université, France \\
  $^2$LIUM, Le Mans Université, France \\
  $^3$LIA, Avignon Université, France \\
  \texttt{antoine.tholly@univ-nantes.fr, jane.wottawa@univ-lemans.fr,} \\
  \texttt{mickael.rouvier@univ-avignon.fr, richard.dufour@univ-nantes.fr}}

\begin{document}
\maketitle
\begin{abstract}
Automatic Speech Recognition (ASR) transcription errors are commonly assessed using metrics that compare them with a reference transcription, such as Word Error Rate (WER), which measures spelling deviations from the reference, or semantic score-based metrics. However, these approaches often overlook what is understandable to humans when interpreting transcription errors. To address this limitation, a new evaluation is proposed that categorizes errors into four levels of severity, further divided into subtypes, based on objective linguistic criteria, contextual patterns, and the use of content words as the unit of analysis. This metric is applied to a benchmark of 10 state-of-the-art ASR systems on French language, encompassing both HMM-based and end-to-end models. Our findings reveal the strengths and weaknesses of each system, identifying those that provide the most comfortable reading experience for users.
\end{abstract}

\section{Introduction}

Every linguistic description is based on theoretical assumptions, whether acknowledged or not. This study is influenced by contextual linguistics~\cite{rastier2015interpretative,col2012gestalt}, in which the meaning of a word is interpreted not only by its form (spelling/morphology, \emph{e.g.}, the five letters of ``plane''), but also by the other words in its semantic network (\emph{i.e.}, words belonging to the same categories and/or the same sequence). The human recognition process considers the surrounding grammatical and lexical/thematic \emph{context} in which the word – whether correct or erroneous – is used; even more so if the word is erroneous.

In the context of Automatic Speech Recognition (ASR), there are two aspects to the interpretation of transcription errors: \emph{detection} (\emph{e.g.}, ``pla'' is an incorrect spelling) and \emph{resolution} (\emph{e.g.}, the reference ``plane'' might be reconstructed from ``pla''). Consequently, we propose that ASR error detection and resolution should be analyzed \emph{within the context of their sentences}. This relationship is bidirectional: an error can \emph{import its solution} from the context (\emph{e.g.}, ``pla'' understood as ``plane'' within the aeronautical theme of a sentence), and it can \emph{export its problem} to the context (e.g., blurring or altering the overall meaning of the sentence).

Traditional error metrics such as Word Error Rate (WER) tend to ignore context. Objectively, a transcription error can be defined as a deviation from the reference. However, this perspective does not account for how users interpret errors solely through the transcription. In the \emph{detection phase}, spelling issues (\emph{e.g.}, ``a pla'' instead of ``a plane'') are seen as errors in all contexts, unlike context-dependent semantic errors (\emph{e.g.}, ``the plate lands on the tarmac''). The same applies to the \emph{resolution phase} of errors: context may or may not be needed to retrieve the reference, and if needed, it may be adequate or insufficient. Furthermore, errors also have various \textit{ consequences} for the understanding of the contextual sequence (\emph{e.g.}, an easily detectable and resolvable error like "choocolate" does not significantly affect other parts of the sentence).

In this paper, we propose degrees of \emph{error severity} from an interpretative perspective in French, alongside an original study of textual automatic transcription errors from various ASR architectures. Section 2 introduces four classes of errors that reflect the severity of errors from the user’s perspective. Errors can have minimal impact when the word is immediately recognizable, such as minor spelling issues without semantic consequences (Section 2.1). Grammatical errors on content words, while not impeding comprehension, contravene established social norms (Section 2.2). Moderate to significant errors require additional effort, as they can only be understood through contextual processing, which may be challenging (Section 2.3). The final type of error is critical, as it completely disrupts understanding due to ambiguity, contextually uninterpretable words or  deletions, or undetectable errors when the mistake matches the sentence meaning and syntactic structure (Section 2.4). Boundary cases and individual variations are also addressed (Section 2.5). Section 3 outlines the protocol using a French corpus and 10 ASR systems differing in modularity, self-supervised learning (SSL) audio models, tokenizers, and training data volume. These systems are benchmarked in Section 4 based on the four error categories, leading to a discussion of the system's capabilities. Section 5 concludes and suggests future perspectives.

\section{A User-based Metric}
\label{sec:apptypology}

The metric aims to rank systems based on their ability to prevent different types of errors, defined according to their perceived severity for humans. This human-centered interpretative criterion contrasts with the purely formal, spelling-deviation criteria commonly employed by metrics such as WER and CER (Character Error Rate). These formal metrics are limited to detecting errors and offer no insight into their semantic distance from the reference—that is, the interpretative effort required to resolve them, if resolution is possible.

Expanding on these limitations, semantic metrics measure the semantic distance between the hypothesis and the reference \cite{zhang2019bertscore,kim2021semantic}, but do not examine the underlying causes of this distance. Considering these causes, we argue that this distance is smallest when context is \textit{superfluous} for error interpretation, greater when context is \textit{necessary}, and greatest when context is \textit{insufficient} or \textit{misleading}.

This semantic distinction also guided the decision to use lexical/content words (nouns, adjectives, verbs, adverbs) as the unit of measurement within the metric. These words provide a simple, homogeneous criterion for objectivity in measurement, while carrying significant information at the word level, essential for interpretative relevance. Although other linguistic units, such as grammatical/function words and higher-level structures like phrases and clauses, are important, combining all linguistic levels where errors occur would not result in a comprehensive metric or clear interpretative criteria. Additionally, grammatical constructions are partially reflected in lexical word errors—first in their inflectional parts, and second, when phrases or clauses are severely distorted, as the core lexical words are interpreted through the distortions they reflect.

The proposed four-category error system is designed to account for all lexical word errors in the corpus, ensuring comprehensive, broad, and objective coverage while minimizing ambiguity during categorization (cf. Section 2.5). These primary categories are further divided into detailed subcategories, which are outlined and exemplified in their respective sections.

To illustrate distinctions between error types and subtypes, variations of a single example, 'gorilla,' are used as a pedagogical tool to facilitate focused comparison. Appendix A provides a classified sample of errors from the French corpus, accompanied by their English translations.

The four error types forming the metric will be revisited in Section 4 (Results) to highlight their insights into the performance of the benchmarked ASR systems.

\subsection{Minimal Impact: Immediate Word Recognition (Lex)}

The least severe category (comprehension-based) regroups cases where the identity of the word is successfully recognized without relying on context (\textit{gorila}, \textit{patato}, \textit{advennture}). 

In the 'Lexical' category, error occurs in the stem part of the word, i.e. without additional issues in the inflectional part (e.g., "two gorila" is to be classified in the 'Grammatical' category). This is similar to how speakers may recognize words they don't know from another related language, such as an English speaker understanding the French lexeme 'compétitivité'. 

A variation of this error involves internal segmentation (or splitting) issues, where the word remains recognizable in isolation, such as "a gor illa" or "a go rilla".

As a clarification, although the context is not required for the solution, it still plays a role in word interpretation (as it always does), particularly because the immediate solution, out-of-context, could still be incorrect (in the case of a 'gorila', the reference would generally be 'gorilla' but could still be another word, like 'guerilla'. Cf. Section 2.4).

\subsection{Special Disruption: Grammatical Botheration (Gram)}

As with the previous type of error, the identity of the word is easily recognized without relying on context, but it contains an error in the inflection, e.g., 'one gorillas.' In the French corpus, this pertains to gender and plural markers on nouns and adjectives, as well as tense and person markers on verbs.

This category reflects the specific role of grammar in both ASR processing and the 'botheration' reactions of end-users. 'Botheration' is a concept from  psycholinguistics, referring to errors that do not affect comprehension but violate established norms and conventions. ~\cite{boettger2018analyzing,smith2015botheration}.

\subsection{Moderate to Significant Difficulty: Effort Requirements from Context Processing (Cotx)}

Erroneous words can be resolved fully or partially with the help of contextual words or structures, but this process requires greater cognitive effort, often at the expense of the end-user's reading comfort. Three subtypes of solution recognition are outlined below.

Local group recognition involves identifying multiword expressions and named entities (e.g., "magilla gor" for the "Magilla Gorilla" cartoon character), as well as contiguous collocations (e.g., "a goril sanctuary").

Broader context recognition is triggered by lexical/thematic cues, such as 'my favorite animal is the gori', or by grammatical triggers like anaphora, as in 'there's a gorilla, and this gril [...]'.

Partial recognition occurs when context provides limited information, such as identifying an animal ('my favorite animal is the gr') or an agent-like entity ('I talked to this gr'). Partial understanding often edges toward the critical error category.

\subsection{Critical Miscommunication: Non-Understandable Errors (Fail)}

The inability to retrieve the solution represents the most severe class of errors. 

Hesitation or ambiguity arises when multiple solutions could apply, as in "[\textit{no additional context}] a gril is an animal" where it’s unclear whether the term refers to 'grizzly' or 'gorilla'. A second case occurs when a unique contemplated solution seems uncertain: the user recognizes 'gorilla' from a spelling error, but is not sure if he can rely on his interpretation. 

The user can also acknowledge an impossible interpretation. The recognition of an impasse in the resolution process can be triggered by several factors: i) uninterpretable word-forms, even with context, such as "there is a gorallo", ii) certain ungrammatical sentences, e.g. "there is a * that" (a missing word is inferred from the incomplete noun phrase), and iii) certain meaningless associations between the context and a (seemingly) correctly spelled word (e.g., "the polarization of semiconductor gorilla"): while this error is detected through lexical semantics incompatibility, no solution is identified.

Finally, misleading interpretations (or false positives) occur when the error is undetectable, such as i) a lexical substitution that fits the context, like "guerilla" instead of "gorilla" in "a guerilla in the forest", or ii) a deletion of a syntactically optional element, as in "I'm in the forest", where the optional adjunct "with a gorilla" is omitted and cannot be perceived.

\subsection{Remarks: Boundary Cases and Individual Variations}

Our categories are clearly delineated for the purposes of the upcoming quantification, grouping the variety of subtypes around a central criterion. However, there is undoubtedly a continuum between errors that can be resolved through context ('Cotx') and those that cannot ('Fail'), and we have already mentioned that partial resolutions often lean toward critical errors. In practice, we do not believe that errors at the edges of categories are fundamentally a problem, for two reasons: firstly, in the distribution of errors into their respective categories (intra-system quantification of errors), they represent infrequent occurrences; and secondly, in the task of benchmarking (inter-system quantification of errors), they primarily need to be addressed consistently, i.e., in the same way across all systems.

Additionally, there are certainly individual variations in resolution abilities, which this typology and the forthcoming quantification appear to overlook. Here again, a clear response can be provided by refining the definitions. Context-resolvable errors ('Cotx') are expected to be resolved by most readers, whereas context-irresolvable errors ('Fail') are expected to remain unresolved by most readers (with the rarer boundary cases likely dividing interpretations). As our categories ground human interpretation on objective linguistic cues, we anticipate that a clear majority of human interpretations will align with these objective linguistic cues. Furthermore, variations between individuals are expected to be similar across systems, rendering them neutral with respect to the benchmark.

\section{Experimental protocol}
\label{sec:experimental_protocol}

The transcription corpus is derived from the REPERE corpus, which consists of French-language French television programs on news events, such as politics and culture.~\cite{giraudel2012repere}. The phonostyle~\cite{de2014qu} can be described as public speaking by communication professionals, in prepared or semi-prepared formats, such as studio interviews and scripted delivery by journalists, and recorded under excellent audio conditions. REPERE had already segmented each broadcast into short audio sequences and expertly transcribed the audio of each sequence (i.e., the reference). We processed these audio segments with 10 different ASR systems to create a corpus of transcriptions. For this study, we used a subset of this corpus, specifically transcriptions of four different broadcasts.

All lexical words in this transcription corpus were classified either as correct (i.e., corresponding to the reference) or as one of the four distinct types of errors described in Section 2. These analyses were conducted by a linguistic expert through Glozz, an annotation tool developed for expert annotation in text corpora~\cite{widlocher2012glozz}. The annotation of the 10 ASR transcriptions from the four broadcasts resulted in a total of 10,007 lexical words (1,125 annotated errors), with overall lexical word accuracy rates ranging from 79.6\% to 93.7\% across the broadcasts (averaged across the 10 systems).

This study evaluated 10 ASR systems from the Kaldi~\cite{povey2011kaldi} and SpeechBrain~\cite{ravanelli2024opensourceconversationalaispeechbrain} toolkits, using various speech recognition methodologies. Two systems are DNN-HMM systems based on Kaldi, while 8 are end-to-end systems from SpeechBrain using various techniques such as SSL (self-supervised learning) audio models or tokenizers (see Table~\ref{table:systems}).

Systems based on Kaldi are prefixed with KD. The KD\_wR and KD\_woR systems used a 3-gram language model for decoding, but KD\_wR also includes an additional posterior rescoring step based on the RNNLM deep neural network language model.

\sloppy Systems based on SpeechBrain are prefixed with SB. The SB\_no\_char, SB\_XLSR\_char, SB\_XLSRFR\_char, SB\_LB1k\_char, SB\_LB3k\_ char, and SB\_LB7k\_char systems use a character tokenizer, while SB\_LB3k\_bpe750 and SB\_LB3k\_bpe1000 a Byte Pair Encoder (BPE) tokenizer. All the systems, except SB\_no\_char, used an SSL audio model. SB\_XLSR\_char and SB\_XLSRFR\_char used the XLS-R model (cross-lingual speech representation based on wav2vec 2.0) whereas the others used the LeBenchmark models (French speech representation based on wav2vec 2.0)~\cite{parcollet2024lebenchmark}. We note that LeBenchmark 1k, 3k, and 7k are pre-trained on 1k, 3k, and 7k hours of unlabeled data, respectively. 

All ASR systems were trained on French data using various corpora (ESTER 1 and 2~\cite{galliano2006corpus,galliano2009ester}, EPAC, ETAPE~\cite{gravier2012etape}, REPERE~\cite{giraudel2012repere}, and internal data), totaling about 940 hours of audio.

\begin{table}[H]
\centering
\resizebox{\columnwidth}{!}{
    \begin{tabular}{ l l l }
	\hline
	  \rowcolor{gray!40} \textbf{Systems}  & \textbf{SSL Audio}  &  \textbf{Tokenizer} \\
	\hline
        SB\_no\_char & No & Character \\
        SB\_XLSR\_char & XLS-R & Character \\
        SB\_XLSRFR\_char & XLS-R FR & Character \\
        SB\_LB1k\_char & LeBenchmark 1k & Character \\
        SB\_LB3k\_char & LeBenchmark 3k & Character \\
        SB\_LB7k\_char & LeBenchmark 7k & Character \\
        SB\_LB3k\_bpe750 & LeBenchmark 3k & BPE 750 \\
        SB\_LB3k\_bpe1000 & LeBenchmark 3k & BPE 1000 \\
	\hline
	\end{tabular}
 }
\caption{Systems overview with different SSL audio models and tokenizers.}
\label{table:systems}
\end{table}

\section{Results}
\label{sec:results}

In total, 1,125 lexical words errors are classified and quantified across 10 systems, along with 8,882 correct lexical words. Each system transcribed an identical corpus. The statistics are organized into 4 categories of errors, described in the metric section (Section 2). 'All' refers to the total of errors (in percentage compared to the total of lexical words). 

Percentages of errors for each system regarding each category are presented in Table 2. The systems are ranked from best to worst on the vertical axis, taking into account the total rate of errors and giving greater weight to Fail errors.

\noindent In Table~\ref{table:results}, we observe that:

\textbf{Immediate Lexical Recognition errors (Lex).} Kaldi systems achieve fewer 'Lex' errors compared to SpeechBrain systems. This can be attributed to the use of a language model, which helps avoid hallucinations of words, unlike the SpeechBrain systems.

\textbf{Grammatical errors (Gram).} The Kaldi system without rescoring achieves the second-lowest score, while the Kaldi system with rescoring achieves one of the best scores. This improvement is due to the RNNLM used as an additional posterior rescoring step, which considers a broader context, allowing for better error correction. Interestingly, LeBenchmark models achieve results equivalent to Kaldi with rescoring, better than the XLS-R model. This indicates  that training an audio SSL model in the target language effectively captures this type of information.

\begin{table}[H]
\centering
\resizebox{\columnwidth}{!}{
%\scriptsize
    \begin{tabular}{ c c c c c c c }
	\hline
	   \textbf{Systems}   & \textbf{(WER)}  &  \textbf{\textit{All}} & \cellcolor{gray!30}{ \textbf{Lex} } & \cellcolor{gray!30}{ \textbf{Gram} } & \cellcolor{gray!30}{ \textbf{Cotx} } & \cellcolor{gray}{ \textbf{Fail} } \\
	\hline
        Total & 18.77 & 11.8 & 2.1 & 2.1 & 2.2 & 5.3 \\
KD\_wR & 13.21 & 5.4 & 0.5 & 1.5 & 0.2 & 3.2  \\
SB\_LB7k\_char & 16.56 & 7.0 & \cellcolor{gray!30}{  2.0 } & 1.3 & 1.6 & 2.2 \\
SB\_LB3k\_bpe750 & 15.33 & 8.4 & \cellcolor{gray!30}{  2.6 } & 1.6 & 1.8 & 2.5 \\
SB\_LB3k\_bpe1000 & 15.98 & 8.5 & \cellcolor{gray!30}{  2.4 } & 1.5 & \cellcolor{gray!30}{  2.2 } & 2.5 \\
SB\_LB3k\_char & \cellcolor{gray!30}{ 17.16 } & 9.5 & \cellcolor{gray}{  3.0 } & \cellcolor{gray!30}{ 2.2 } & \cellcolor{gray!30}{  2.0 } & 2.4 \\
KD\_woR & 15.43 & 7.6  & 0.3 &  \cellcolor{gray}3.1  & 0.3 & \cellcolor{gray!30}{ 3.9 } \\
SB\_LB1k\_char & \cellcolor{gray!30}{ 18.94 } & \cellcolor{gray!30}{  10.8 } & \cellcolor{gray!30}{  2.2 } & \cellcolor{gray!30}{  1.9 } & \cellcolor{gray!30}{ 
 2.1 } & \cellcolor{gray!30}{  4.6 } \\
SB\_XLSR\_char & \cellcolor{gray}{ 22.69 } & \cellcolor{gray}{ 14.9 } & \cellcolor{gray}{ 3.0 } & \cellcolor{gray!30}{ 2.0 } & \cellcolor{gray}{ 4.6 } & \cellcolor{gray!30}{ 5.3 } \\
SB\_XLSRFR\_char & \cellcolor{gray}{ 21.48 } & \cellcolor{gray}{ 16.6 } & \cellcolor{gray}{ 3.6 } & \cellcolor{gray}{ 2.6 } & \cellcolor{gray}{ 4.0 } & \cellcolor{gray!30}{ 6.4 } \\
SB\_no\_char & \cellcolor{gray}{ 30.94 } & \cellcolor{gray}{ 23.4 } & \cellcolor{gray!30}{ 2.2 } & \cellcolor{gray}{ 3.4 } & \cellcolor{gray}{ 3.9 } & \cellcolor{gray}{ 14.0 } \\

	\hline
	\end{tabular}
 }
\caption{Error rates for each ASR system across categories, color-coded: white for lowest errors, light grey for moderate errors, and dark grey for highest errors.}
\label{table:results}
\end{table}

\textbf{Contextual errors (Cotx).} Kaldi systems achieve better results due to their language model. We observe that the LeBenchmark models, trained on a large amount of data, can manage to correct these errors. Additionally, BPE tokenizers yield better results than character tokenizers.

\textbf{Failure errors (Fail).} The best results are obtained with LeBenchmark models using BPE tokenizers, as well as LeBenchmark models with character tokenizers and ample training data. Among these LeBenchmark models with character tokenizers, a notable reduction in the ‘Fail’ error rate is observed as the training data increases from 1K (4.6\% error rate) to 3K (2.4\% error rate). Kaldi models demonstrate moderate performance. Regarding SLL Audio, all LeBenchmark models outperform XLR-S models. The Seq2Seq model without SLL Audio (i.e., SB\_no\_char) performs the worst, as it also did with Gram and Cotx errors and overall: this is consistent with previous observations, as this system does not integrate a language model (and is limited to characters) while having no rich acoustic information.

\textbf{\textit{Closing Analysis.}} The Kaldi system with rescoring achieves the best overall performance, as reflected in the 'All' error rates. However, the LeBenchmark model with character tokenizers and 7K of training data, while second overall, ranks slightly stronger than the former in addressing the most critical errors ('Fail' error rates). It is worth noting that LeBenchmark models with BPE tokenizers also demonstrate commendable performance overall, despite being limited to 3K of training data.

\textbf{\textit{WER comparison}.} To assess these figures, we compare them with the WER results from the entire REPERE corpus, providing a broader evaluation of system error performance. The key comparison is not the absolute error rate (WER rate is higher): grammatical words are excluded from our experiment, and it only suggests that the audio subset used for the experiment was easier to process. The critical metric lies in the deviation among systems: our results and the WER results follow similar trends, offering strong evidence of our method's reliability. More importantly, our study introduces finer-grained dimensions for benchmarking these systems, extending beyond what the automatic metric alone can measure, as outlined above.

\textbf{\textit{Statistical Relevance.}} For this corpus, the percentage difference that is statistically significant  between two error rates has been calculated to be approximately 1.7\%. Between previously discussed system differences, this significance threshold is usually exceeded.

\section{Conclusions and Perspectives}
\label{sec:conclusions}

A new typology of transcription error severity in ASR systems is proposed, based on a model of their reception by end users. This approach allows for both qualitative and quantitative analysis of errors from 10 French ASR systems, highlighting varying capabilities depending on the used architectures. Through this analysis, the method itself proves effective in the ASR benchmarking task.

This original study is of interest for improving ASR performances: the integration of a user’s interpretative model provides valuable feedback, helping align ASR systems more closely with user expectations. Furthermore, the theoretical explanations based on English errors, combined with the performance analysis of French errors, demonstrate the applicability of this benchmarking method to multiple languages.

Looking ahead, severity criteria will be refined through a perception test evaluating how participants perceive ASR errors. This will help correlate audience assessments of error severity with error frequency across systems, leading to a more detailed and comprehensive performance evaluation.

\section{Limitations}

\textbf{Single Expert.} Only one linguistic expert was involved in the annotation process, which may introduce a bias. The interpretation quality was prioritized at this stage, focusing on consistency in human interpretation. The methodology aimed to mirror end-user perspectives to some extent, but further work could involve additional linguistic experts and user-centered validation to enhance objectivity and reliability.

\noindent\textbf{Benchmarking and Data Scope.} Although the corpus contained about 10,000 lexical words and 1,200 errors, only about 120 errors per system were ultimately categorized and benchmarked. This could limit the breadth of the comparison. 

Additionally, finer-grained errors that had been annotated were finally included in larger categories of quantification, due to their scarcity, further constraining the scope of the benchmarking. Future evaluations will benefit from expanding the size of the corpus to increase the number of quantified categories and the statistical relevance of the results.

\section*{Acknowledgements}

This research received funding from the Agence Nationale de la Recherche (ANR), France, as part of the DIETS project (Automatic Diagnosis of Errors in End-to-End Speech Transcription Systems from the User's Perspective).

We also wish to thank Thibault Bañeras-Roux for his informal contribution.

\bibliography{custom}

\newpage

\appendix
\section{Appendix: Sample of Error Types and Subtypes.}
\vspace{0.2cm}
\hspace*{10pt}\textbf{\textit{Formatting of examples.}} Errors are highlighted in \textbf{bold}, followed by their reference in [brackets]. Where applicable, context words serving as resolution cues are \underline{underlined}. Missing words or  morphemes are indicated by an asterisk (*).

\subsection{(LEX) Minimal Impact: Immediate Word Recognition.}

\hspace*{10pt}\textbf{Subtype 1.1: Minor distortion of word stem, preserving comprehension}.

"syndic\textbf{t}ats" [syndicats]  —  (\textit{unions}).

"comp\textbf{a}titivité" [comp\textbf{é}titivité]  —  (\textit{competitiveness}).

\vspace{0.2cm}
\textbf{Subtype 1.2: Word segmentation error, preserving comprehension}.

"la pât\textbf{ie n}oire" [la patinoire]  — (\textit{the ice skating rink}).

"une l\textbf{e ç}on" [une leçon]  — (\textit{a lesson}).

\subsection{(GRAM) Special Disruption: Grammatical Botheration.}

\hspace*{10pt}\textbf{Subtype 2.1: Verbal inflection error}.

\textit{Tense.} "le comité qui a organis\textbf{ai}" [organis\textbf{é}]  — (\textit{the committee that organized}).

\textit{Person.} "qu'il renonçai* à" [renonçai\textbf{t}]  — (\textit{that he gave up on)}.

\vspace{0.2cm}
\textbf{Subtype 2.2: Nominal and adjectival inflection error}.

\textit{Gender.} "au palmarès important\textbf{e}" [important]  — (\textit{with an important track record}).

\textit{Number.} "les papys rockeur*"  [rockeur\textbf{s}]  — (\textit{the rocking grandpa*}[s]).

\subsection{(CTX) Moderate to Significant Difficulty: Effort Requirements from Context Processing.}

\hspace*{10pt}\textbf{Subtype 3.1: Local context resolution: multiword expressions}.

\textit{Compounds.} "une \textbf{nombre} [onde] \underline{de choc}"  — (\textit{a \underline{shock}\textbf{whale}} [shockwave]).

\textit{Contiguous collocations}. "la \underline{viande} \textbf{a lal} [halal]" —  (\textit{\textbf{a lal} }[halal]\textit{ \underline{meat}}).

\textit{Named entities.} "\underline{valérie} \textbf{pécrese}" [Pécre\textbf{s}se] .

\vspace{0.2cm}
\textbf{Subtype 3.2: Broader context resolution: lexical/thematic or syntactic comprehension cues}.

\textit{Lexical relations.} "ressembler à la \textbf{gresse} [Grèce] et à \underline{l'espagne}"  — (\textit{to resemble \textbf{gresse} }[Greece]\textit{ and \underline{spain}}).  — Lexical relation: "Spain" and "Greece" ("gresse") are \textit{coordinate terms} (European countries).

\textit{Lexical properties.} "il y a des \textbf{sars} [stars] mais alors après il y a aussi beaucoup de femmes qui sont beaucoup moins \underline{connues}"  —  (\textit{there are \textbf{sars} }[stars]\textit{, but then after that there are also many women who are much less \underline{famous}}). — Interpretation: "famous" is a \textit{defining characteristics} of "stars" ("sars").

\textit{Lexical fields.} "dites moi euh vous êtes spécialisé dans les \textbf{tracs} [tracts] de l'\underline{ump} parce qu'il y a aussi des \textbf{tracs} [tracts] du \underline{ps} qui \underline{expliquent} l'inverse." —  \textit{(tell me uh you are specialized in \textbf{tracs} }[tracts/leaflets]\textit{ of the \underline{ump} because there are also \textbf{tracs} }[tracts/leaflets]\textit{ of the \underline{ps} that \underline{explain} the opposite)}, UMP and PS being political parties  — Lexical field: \textit{politics}.

\textit{Syntactic structure.} "\underline{monsieur bayrou} \textbf{je vous} [joue] \underline{les prophètes}" — (\textit{\underline{monsieur bayrou} \textbf{I you} }[plays]\textit{ \underline{the prophet}})  — Syntactic cues: the sequence is detected as ungrammatical (1 noun phrase + 2 pronouns + 1 noun phrase). The word order suggests the following resolution: 1 noun phrase subject + 1 verb + 1 noun phrase object, all the more so as the two pronouns "je" (\textipa{/Z@/}) + "vous" (\textipa{/vu/}) bear a phonetic resemblance to the verb "joue" (\textipa{/Zu/}).

\textit{Anaphora.} "pendant cinq ans la droite monsieur fillon et monsieur \underline{sarkozy} monsieur \textbf{skozy} [Sarkozy] d'abord monsieur fillon ensuite"  — (\textit{for five years the right party monsieur fillon and monsieur \underline{sarkozy} monsieur \textbf{skozy} }[Sarkozy]\textit{ first then monsieur fillon}).

\vspace{0.2cm}
\hspace*{10pt}\textbf{Subtype 3.3: Partial context resolution: limited comprehension cues}.

"à l'occasion de la \underline{sortie en salle} de \textbf{france canouni} [Frankenweenie], le réalisateur propose une exposition autour de la création de ce \underline{film d'animation}" —  (\textit{on the occasion of the \underline{theatrical release} of \textbf{france canouni} }[Frankenweenie]\textit{, the director is offering an exhibition about the creation of this \underline{animated film}})  — Partial interpretation: the error refers to the name of an animated film just released, but its identity remains unclear.

\subsection{(FAIL) Critical Miscommunication: Non-Understandable Errors.}

\hspace*{10pt}\textbf{Subtype 4.1: Ambiguity}.

\textit{Uncertainty between competing solutions}. "on vient de nous expliquer que trois milliards serait un \textbf{denjeu} [enjeu] national non ça n'est pas sérieux" — (\textit{we were just told that three billion would be a national \textbf{danjeu} no that's not serious}) — Ambiguous interpretation: does "denjeu" [\textipa{d\~{a}j\o}] refer to "danger" [\textipa{d\~{a}je}] (\textit{danger}) or to "enjeu" [\textipa{\~{a}j\o}] (\textit{stake})?

\textit{Hesitation to accept a considered solution}. "je \textbf{re guemarque} [remarque] là-dedans que pour revenir à l'équilibre [...]"  — \textit{I \textbf{nog otice} }[notice]\textit{ in this that to return to balance }[...]) — Ambiguous interpretation: could "I nog otice" mean "I notice"?

\vspace{0.2cm}
\hspace*{10pt}\textbf{Subtype 4.2: Acknowledged impossible interpretation}.

\textit{A detected but unsolvable form distortion}. "un univers \textbf{wason} [foisonnant] que les Parisiens pourront découvrir" — \textit{(a \textbf{tymnge} }[teeming]\textit{ universe that Parisians will be able to discover)}.

\textit{A detected but unsolvable lexical incompatibility with contextual meaning}. "ceux qui ont créé le désastre brandissent l'épouvantail \textbf{lagos} [de la gauche]" — (those who created the disaster are brandishing the \textbf{lagos} looming threat \textit{[the looming threat of the Left]}).

\textit{A detected but unsolvable syntactic deletion}. "qu'un premier ministre * [dise] si la gauche passe la zone euro va s'effondrer" — (\textit{that a prime minister * }[would say]\textit{ that if the left comes to power the eurozone will collapse}), "dise"/"would say" being a missing word.

\vspace{0.2cm}
\hspace*{10pt}\textbf{Subtype 4.3: Misleading interpretation (false positives)}.

\textit{Undetectable lexical substitution}. "on va nous réunir à treize heures trente de la sauvette pour présenter le \textbf{problème} [programme] de stabilité de la France" — (\textit{we'll be meeting at one thirty in a rush to present France's stability \textbf{problem} }[program]).

\textit{Undetectable deletion of an optional syntactic element.} " * [merci] Dominique de Montvalon et alors France Soir peut-être pourrait renaître de ses cendres on verra" — ( * [thank you] \textit{Dominique de Montvalon and then France Soir might rise from its ashes we'll see}).

\end{document}